\documentclass[lettersize,journal]{IEEEtran}
\usepackage{amsmath,amsfonts}
\usepackage{algorithmic}
\usepackage{algorithm}
\usepackage{array}
\usepackage[caption=false,font=normalsize,labelfont=sf,textfont=sf]{subfig}
\usepackage{textcomp}
\usepackage{stfloats}
\usepackage{url}
\usepackage{verbatim}
\usepackage{graphicx}
\usepackage{cite}
\usepackage{hyperref}

\usepackage[para]{threeparttable}
\usepackage{multirow}
\usepackage{graphicx}
\usepackage{amssymb}
\usepackage{booktabs}
\begin{document}

\title{Fine-grained Action Analysis: A Multi-modality and Multi-task Dataset of Figure Skating}

\author{Sheng-Lan Liu, Yu-Ning Ding, Gang Yan, Si-Fan Zhang, Jin-Rong Zhang, Wen-Yue Chen, Ning Zhou, Xue-Hai Xu, Hao Liu
\thanks{Sheng-Lan Liu, Yu-Ning Ding, Gang Yan, Si-Fan Zhang, Jin-Rong Zhang, Wen-Yue Chen, Ning Zhou, Xue-Hai Xu and Hao Liu are with the Computer Science and Technology, Dalian University of Technology, Dalian 116024, China. E-mail: (liusl@dlut.edu.cn; \{rookie233, yaner, 201981131, zjr15272565639, 20121212\} @mail.dlut.edu.cn); zhouyuxuan98@gmail.com;  3348530532@mail.dlut.edu.cn; 610216579@qq.com}
\thanks{(Corresponding author: Sheng-Lan Liu.)}}

\markboth{Journal of \LaTeX\ Class Files,~Vol.~14, No.~8, August~2021}%
{Shell \MakeLowercase{\textit{et al.}}: A Sample Article Using IEEEtran.cls for IEEE Journals}


\maketitle

\begin{abstract}
The fine-grained action analysis of the existing action datasets is challenged by insufficient action categories, low fine granularities, limited modalities, and tasks. In this paper, we propose a multi-modality and multi-task dataset of Figure Skating (MMFS) which was collected from the World Figure Skating Championships. MMFS, which possesses action recognition and action quality assessment, captures RGB, skeleton, and is collected from the score of actions from 11671 clips with 256 categories including spatial and temporal labels. The key contributions of our dataset fall into three aspects as follows. (1) Independently spatial and temporal categories are first proposed to further explore fine-grained action recognition and quality assessment. (2) MMFS first introduces the skeleton modality for complex fine-grained action quality assessment. (3) Our multi-modality and multi-task dataset encourages more action analysis models. To benchmark our dataset, we adopt RGB-based and skeleton-based baseline methods for action recognition and action quality assessment. Our dataset is publicly available at \href{https://github.com/dingyn-Reno/MMFS/tree/main}{https://github.com/dingyn-Reno/MMFS/tree/main}.
\end{abstract}


\section{Introduction}
\label{sec:intro}

With the deeper exploration in action recognition, fine-grained human action recognition has long been a question of great interest in a wide range of fields\cite{RN156}\cite{RN183}. The content of videos with fine-grained human action is composed of different combinations of scenes, tools (fixed or non-fixed), objects (dynamic or static), and persons. In recent years, the motion-centered fine-grained action recognition datasets such as \cite{RN105}\cite{diving48}\cite{RN122}, have paid more attention to creating new action categories with the combinations of tools and human actions\cite{RN154}. Recent developments in fine-grained human action recognition have heightened the need for professional sports. Compared with the existing datasets with different scenes, professional sport is challenging because human action will play an important role in a single scene\cite{RN105}\cite{diving48}. Meanwhile, the size of our dataset and the number of action categories are untouchable by the combination of human action and non-fixed tools (More details will be elaborated in Sec.2.). Therefore, it is easier to show more details of fine-grained actions with non-fixed tools in a single scene. The challenges of fine-grained human action datasets are mainly derived from 1) Annotation quality and 2) Impact of $pv$ (pose variation) and $tv$ (temporal action variation) on $cl$ (change of label). It's worth noting that $tv$ is influenced by the number of repeated action units and the speed variation among actions (one or both will be represented in an action sequence). Such impact can be denoted as $P(cl|pv)$(or $P(cl|tv)$), in which $P$ indicates the probability of label changing under the condition of $pv$ or $tv$. The reader should bear in mind that the fine-grained action is based on small inter-class variance. We can divide the fine-grained action into fine-grained semantics and fine-grained complexity. Given the above, the disadvantages of the existing datasets can be listed as follows:


\textbf{Fine-grained semantics.} The fine-grained semantics that can be simply described as $P(cl|pv)\rightarrow1$ and $P(cl|tv)\rightarrow1$ will lead to small intra-class variance. The fine-grained motion-centered action datasets place more emphasis on the quality of action annotation (requires professionalism and expert participation), the number of categories, and temporal fine-grained semantics \cite{RN168}. Owing to the lack of official document or real-time labeling by experts, most datasets (e.g. dance\cite{RN181}, Taichi\cite{RN177}, etc) are weak in labeling, the accuracy and professionalism of labels are limited \cite{RN171}. Moreover, restricted by fixed tools (e.g. pommel horse in FineGym\cite{RN105}) or strategic objects (e.g. basketball\cite{RN182}), the number of fine-grained categories in the existing human action datasets is insufficient (see Tab. \ref{tab:tab1}). In fact, the relationship between $pv$ and $cl$ tends to be formulated by $P(cl|pv)\rightarrow1$, which means the larger $pv$ is, the more the number of categories will be. And this is also what most of the existing datasets adopt to increase the number of fine-grained categories. Yet, $tv$ (temporal action variation), which also contributes to ensuring categories, quite goes by the board. That is, the condition $P(cl|tv)\rightarrow1$ is rarely met so that the fine granularity would not increase at the temporal level.

\textbf{Fine-grained complexity.} The fine-grained complexity is mainly reflected in two aspects: 1) the large duration and speed variance 2) $P(cl|pv)\rightarrow0$ and $P(cl|tv)\rightarrow0$. Action categories that only contain fine-grained complexity without fine-grained semantics will lead to large intra-class variance. Up to now, most studies in the field of human action datasets\cite{RN154} have only focused on fine-grained semantics and limited spatial fine-grained complexity (See Fig. \ref{fig:fig1}). There has been no detailed investigation of fine-grained complexity about temporal levels\cite{RN168} and spatio-temporal levels. For the existing recognition models, it is less challenging to obtain well-trained models from the existing fine-grained human action datasets in the case of the complex spatio-temporal features of fine-grained complexity is inadequate.

\IEEEpubidadjcol

\textbf{Modality.} There only exist RGB and flow features for most existing fine-grained human action datasets. It is unfortunate that the skeleton features in FineGym dataset \cite{RN105}, which consists of RGB, flow, and skeleton features simultaneously, is exacted incompletely. Accordingly, the development of Fine-grained skeleton-based models is limited in the field of human action recognition.

Taken together, reliable action labels are expected to ensure that the change of label ($cl$) impacted by $tv$ and $pv$ is accurate. The number of fine-grained actions and the intra-class variance are limited. A small action dataset FSD-10 \cite{liu2020fsd} is proposed for fine-grained action analysis with the above characteristics but without independent spatial/temporal fine-grained semantics and large-scale samples. We thus propose a new figure skating dataset named MMFS (Multi-modality Multi-task dataset of Figure Skating), collected from videos with high definition (720P) in the World Figure Skating Championships. Compared with the existing human action datasets, the advantages of MMFS can be summarized as follows:

\begin{figure}
   \centering
   \includegraphics[width=0.5\textwidth]{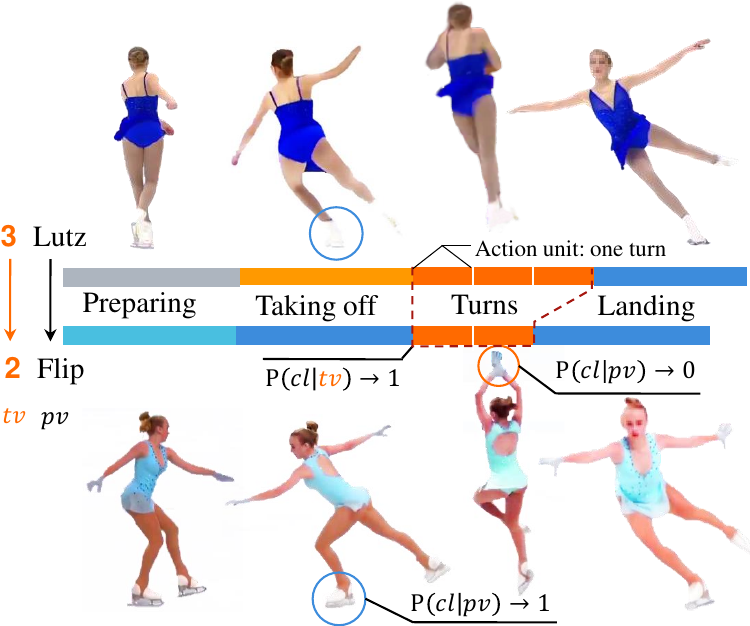}
   \caption{Examples of spatio-temporal fine-grained action categories. Spatially, Lutz and Flip can be classified by $P(cl|pv)\rightarrow1$. Raising a hand in 2Flip will not change the label, which indicates $P(cl|pv)\rightarrow0$. Temporally, $P(cl|tv)\rightarrow1$ denotes different turns that will change the action label.}
   \label{fig:fig1}
 \end{figure}

\textbf{Strong annotation.} Weak annotation is labeled by trained people. Medium annotation is indexed by trained people and official documents. Strong annotation is annotated by experts and an official document, which means MMFS is jointly annotated by both real-time expert determination and proficient annotators under the help of an official document, which can be used to guarantee the label is equipped with accuracy and professionalism.

\textbf{Independently Spatial Label (SL) and Temporal Label (TL).} MMFS dataset has \textit{spatio-temporal fine-grained semantics:} Skates, as wearable and non-fixed tools, assist body movements to add richer pose details to actions \cite{RN154}, introducing more complex spatial fine-grained actions. The number of fine-grained actions will be increased by $tv$ and $pv$ as part of action units change for one given action (please see Fig. \ref{fig:fig1} for details). To further research action recognition at both spatial and temporal levels, we propose integrally spatial and temporal labels in MMFS. Note that the prediction of temporal labels is more difficult than spatial ones. Temporal semantics indicates more rigorous requirements than spatial semantics because the large duration and speed variance lead to the large intra-class variance.  A hierarchical label structure including temporal and spatial labels is built to compare the fine-grained spatial and temporal semantics.

\textbf{High complexity of spatio-temporal fine-grained action categories.} 1) In comparison with the other datasets, the large duration and speed variance of actions make temporal granularity could be adequately demonstrated. For instance, the Jump could be completed within 2s, while the StepSequence would last from 12s to 68s. The longer average duration of MMFS indicates that more action units can be included in action (see Fig. \ref{fig:fig1}). 2) There are sufficient cases of $P(cl|pv)\rightarrow0$ and $P(cl|tv)\rightarrow0$ in our dataset. More action units and complex spatio-temporal features can maintain the large intra-class variance of fine-grained actions, even with the increasing number of fine-grained action categories (see Section \ref{sec:section3} for details).

\textbf{Multi-modality.} In addition to the RGB feature, the MMFS dataset has the full-body skeleton feature, which offers a great challenge to design remarkable multi-modality models.

\textbf{Multi-task.} MMFS, which includes action recognition and action quality assessment tasks, is currently the largest multi-modality action quality assessment dataset. The score of skating is determined by the quality of the movement and the rules of the International Skating Union (ISU). To be specific, the score of each movement is composed of basic value (BV) and grade of execution (GOE). Therefore, the scoring system is relatively complex, which brings greater challenges to the scoring model.

According to the characteristics and challenges of MMFS, extensive experiments are conducted, including state-of-the-art RGB-based and skeleton-based action recognition models with different input modalities (RGB, flow, and skeleton features). The experiments indicate that: 1) The duration and speed variance of the dataset is large, which makes it difficult to recognize tv-dominated actions; 2) The accuracy of semantic fine-grained actions could be more easily enhanced than that of fine-grained complex ($P(cl|pv)\rightarrow0$ or $P(cl|tv)\rightarrow0$) actions by increasing the number of input frames.

Overall, this work contributes to the fine-grained action field in two aspects:

(1) To our best knowledge, MMFS is the first fine-grained action dataset with strong annotation, high fine-grained spatio-temporal complexity, multi-modality, and multi-task characteristics.

(2) MMFS is challenging to the existing state-of-the-art action recognition models. The dataset can be utilized to exploit more excellent models for action-related tasks, provides inspiration for future exploration in this field.

MMFS involves fine-grained action recognition and action quality assessment tasks. According to the characteristics of MMFS, extensive experiments are conducted, including mainstream RGB-based and skeleton-based action recognition models with different input modalities (RGB and skeleton features). The experiments indicate the challenges of our benchmark, which highlights the need for further research on fine-grained action analysis.
\section{Related Work}
\begin{table*}[]
    \centering
	 \caption{A summary of existing action datasets.}
	
	 \label{tab:tab1}
    \begin{threeparttable}
    \setlength{\tabcolsep}{0.7mm}{
    \resizebox{\textwidth}{!}{
        \begin{tabular}{c|cccccccccccc}
            \hline
                                                & \multicolumn{1}{c|}{Coarse-grained} & \multicolumn{1}{c|}{Skeleton} & \multicolumn{6}{c|}{Fine-grained AR Datasets}                                                                                       & \multicolumn{3}{c|}{Fine-grained AQA Datasets}                        & Ours              \\
                                                & \multicolumn{1}{c|}{Kinetics}       & \multicolumn{1}{c|}{NTU}      & TaiChi                & Diving48   & FSD-10       & Basketball        & FineGym           & \multicolumn{1}{c|}{Muti-sport}               & AQA-7      &   MTL-AQA    & \multicolumn{1}{c|}{FineDiving} & MMFS              \\ \hline
            Years                               & 2017                                & 2019                          & 2017                         & 2018      &  2020       & 2020              & 2020              & 2021                            & 2019              & 2019                         & 2022     & 2023         \\ \hline
            RGB/Flow                            & $\checkmark     $                   & $\checkmark     $             & $\checkmark     $ & $\checkmark     $ & $\checkmark     $ & $\checkmark     $ & $\checkmark     $ & $\checkmark     $               & $\checkmark     $ & $\checkmark     $  & $\checkmark     $  & $\checkmark     $ \\ \hline
            Skeleton                            & $\times        $                    & $\checkmark     $             & $\times        $  & $\times       $   & $\checkmark     $   & $\times       $   & $\checkmark  $   & $\times       $        & $\times       $   & $\times       $    & $\times        $   & $\checkmark     $ \\ \hline
            Fine-grained AR                     & $\times        $                    & $\checkmark     $             & $\checkmark     $ & $\checkmark     $ & $\checkmark     $ & $\checkmark     $ & $\checkmark     $ & $\checkmark     $               & $\times        $ & $\checkmark     $ & $\times        $    & $\checkmark     $ \\ \hline
            Fine-grained AQA                     & $\times        $                    & $\times        $              & $\times        $  & $\times        $  & $\checkmark\tnote{a}$   & $\times        $  & $\times        $  & $\times        $                & $\checkmark     $ & $\checkmark     $    & $\checkmark     $        & $\checkmark     $ \\ \hline
            Single-sport                        & $\times        $                    & $\times        $              & $\checkmark     $ & $\checkmark     $ & $\checkmark     $ & $\checkmark     $ & $\times        $  & $\times        $                & $\checkmark     $ & $\checkmark     $   & $\checkmark     $         & $\checkmark     $ \\ \hline
            \multicolumn{1}{c|}{SL/TL} & -           & -                           & -               & N/A\tnote{b}          & N/A      & -               & N/A\tnote{b}      & -         &  N/A\tnote{b}           & N/A\tnote{b}     & N/A\tnote{b}    & 24/22                \\ \hline
            Annotation                          & -                                   & -                             & Medium            & Strong            & Strong            & Weak              & Medium            & Medium                          & Strong            & Strong          & Strong             & Strong            \\ \hline
            Classes                             & 600                                 & 120                           & 58                    & 48            & 10         & 26                & 530               & 66                              & N/A\tnote{c}                 & 16              & 52             & 256               \\ \hline
            Clips                               & 500000                              & 114480                        & 2772               & 18404   & 1484          & 3399              & 4883\tnote{d}              & 3200                            &  1106              & 1412         &  3000              & 11671            \\ \hline
            \end{tabular}}
    }
    \begin{tablenotes}
        \footnotesize
        \item[a] The dataset offers only the scores without experimental results for the action quality assessment task.\quad\\
        \item[b] Part of actions have a temporal label, but are not illustrated separately.
    \end{tablenotes}
    \begin{tablenotes}
        \footnotesize
        \item[c] The dataset only has action quality assessment task.\quad\quad\quad\quad\quad\quad
        \item[d] The experiments of action recognition are conducted on Element-level.
    \end{tablenotes}
    \end{threeparttable}

 \end{table*}

\textbf{Coarse-grained Action Recognition Dataset.} Coarse-grained datasets always focus on the combination of multiple content elements of videos, such as HMDB51 \cite{RN173}, UCF101 \cite{soomro2012ucf101} and ActivityNet \cite{caba2015activitynet} (and also include large scale datasets something-something \cite{RN126}, Kinetics \cite{RN179}, Moments \cite{RN103} and AViD \cite{piergiovanni2020avid}). The discrimination of these datasets relies on elements (scenes, objects, or tools) rather than the person \cite{lyu2020identity}. In order to focus on the motion of video datasets, motion-centered research began to attract more attention. KTH \cite{RN176} and Weizmann \cite{gorelick2007actions} are early coarse-grained motion recognition datasets without background interference. To enhance the quality of the motion in the dataset, professional sports datasets are involved for high-level human motion expression, such as UCF sport \cite{rodriguez2008action} and Sport-1M \cite{karpathy2014large}, which enhances the number of categories and the variance of action. However, the coarse-grained datasets can not be used to develop fine-grained action analysis models of sports.

 \textbf{Fine-grained Video Dataset.} To weaken the category discriminability of scene and object \cite{RN185} and to deepen understanding of videos, researchers focus more on fine-grained action recognition (AR) datasets. Many simple sports based on balls (like football \cite{RN133}, basketball \cite{RN182}) and body (such as Tai Chi \cite{RN177} and Karate \cite{RN115}) without complex rules are presented to facilitate fine-grained action dataset. Then, more complex sports datasets like MIT-skating \cite{AQA}, diving48 \cite{diving48}, FSD-10 \cite{liu2020fsd} and FineGym \cite{RN105} are proposed to further explore the video understanding. However, these mentioned fine-grained datasets above can not be employed to promote multi-modality and multi-task models.

 \textbf{Multi-modality, Multi-task Dataset.} Some fine-grained datasets are presented to generate multi-task models like MultiSports \cite{MultiSports} (Spatio-Temporal action detection). Moreover, many datasets (such as AQA \cite{AQA} AQA-7 \cite{AQA-7} and FineDiving \cite{xu2022finediving}) are coming up for action quality assessment, where MTL-AQA\cite{MTL-AQA} proposes a multi-task model to process action quality assessment (AQA) and action recognition. MTL-AQA is a diving dataset, but it provides limited fine-grained types (all action types are combinations of a small number of actions). Besides, the pose of action is of great concern in the AQA task, which can distinguish the key to action changes. Yet the skeleton modality only applies to action recognition like NTU \cite{RN119}. In comparison, the size of MMFS is larger than MTL-AQA's and an extra modality can be utilized for action quality assessment. Besides, the data and experiments on the temporal label are rarely mentioned in previous research work. The specific comparison of related datasets is listed in Tab. \ref{tab:tab1}.
\section{Dataset}\label{sec:section3}

MMFS, a multi-task and multi-modality dataset, is challenging for fine-grained action analysis. In this section, the construction of the MMFS dataset is introduced in detail, including data preparation, data annotation, and quality control. Then, we demonstrate the statistical properties and challenges of MMFS.

\subsection{Dataset Construction}

\textbf{Data Preparation.} We collect 107 competition videos of the World Figure Skating Championships from 2017 to 2019 as original videos which are standardized to 30fps with high resolutions on Youtube (720p). Then, the videos are segmented according to 439 figure skaters of two individual items (men, ladies). Each segmented pre-cut video is a complete performance of one skater for checking fine-grained action annotation results and training annotators.

\begin{figure}[h]
    \centering
    \includegraphics[width=0.8\linewidth]{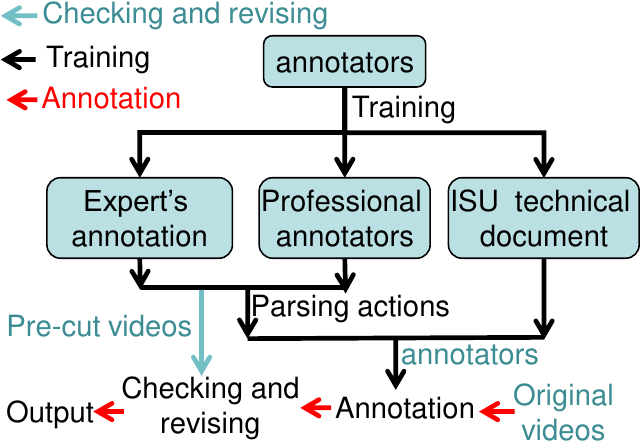}
    \caption{The process of strong annotation.}
    \label{fig:fig2}
\end{figure}

\textbf{Data Annotation.} We annotate two semantics levels for the MMFS dataset, including 3 sets and 256 fine-grained categories(more details of 256 categories of MMFS could be found on our project page). Before annotating the original videos, all the annotators had been trained by professional annotators with figure skating knowledge combining experts' annotation information of all sampled actions in pre-cut videos. From experts' annotation to proficient annotators parsing, combining ISU technical documents is a new strong annotation structure that is an assurance for annotation of MMFS. The official document is referenced by (proficient) annotators during all annotation procedures. The main steps of annotation can be summarized as follows (see Fig. \ref{fig:fig2}). First, the start to the end frames of one action (as a clip) in the original videos are determined according to the provided experts' ground truth in the original videos (see Fig. \ref{fig:fig5}). Then, the incomplete and redundant clips of the original videos have been removed before annotation. At last, all the clips will be annotated manually.

\textbf{Quality Control.} In order to ensure the quality of the MMFS, we adopt the following control methods. 1) Before the formal annotation task, the annotators are evaluated to be competent in this annotation work. 2)It is the key to ensure annotation quality by the information board in the upper left corner of videos, which can not only assist in editing videos but also provide GroundTruth for clips. 3) Professional annotators check and revise all the annotations of actions by leveraging pre-cut videos and all the clips of original videos.

\subsection{Dataset Statistics}
\begin{figure}
   \centering
   \includegraphics[width=1\linewidth]{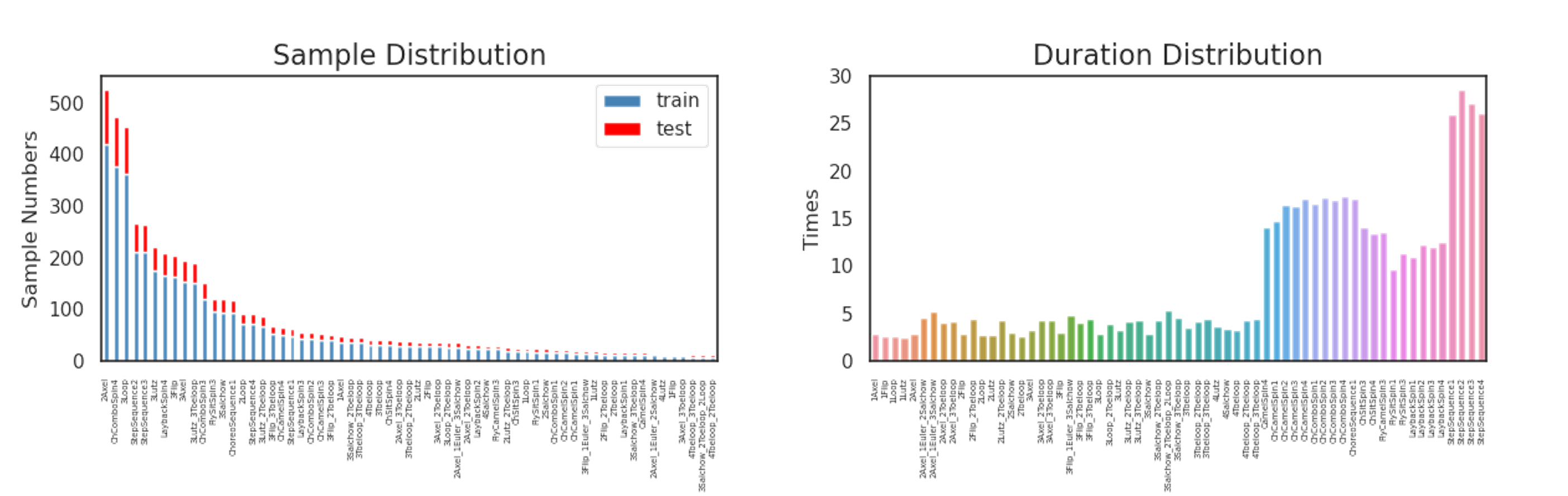}
   \caption{{(}a{)} Samples distribution {(}b{)} Mean duration distribution}
   \label{fig:fig3}
\end{figure}

MMFS contains 11671 clips captured from 107 competition videos, totaling 35.38 hours. To balance the sample distribution of MMFS, we select 63 categories out of 256 categories by filtering insufficient data. Finally, 5104 samples are selected to construct MMFS-63. The samples of the training set and the test set show the characteristics of Heavy-tailed distribution in MMFS-63 (see Fig. \ref{fig:fig3}). The average duration of each category is shown as Fig. \ref{fig:fig3}(b). Specifically, the total video duration of the selected samples reaches 16.35h and the average duration is 11.54s. The duration ranges of actions are from 0.83s to 84.53s with a standard deviation of 10.11s. Compared with the existing datasets \cite{RN105}\cite{diving48}, in MMFS-63, the average duration is longer and the variance of duration is larger, so more fine-grained related properties can be obtained to bring more challenges.

\subsection{Dataset Characteristics}
\begin{figure}
    \centering
    \includegraphics[width=1\linewidth]{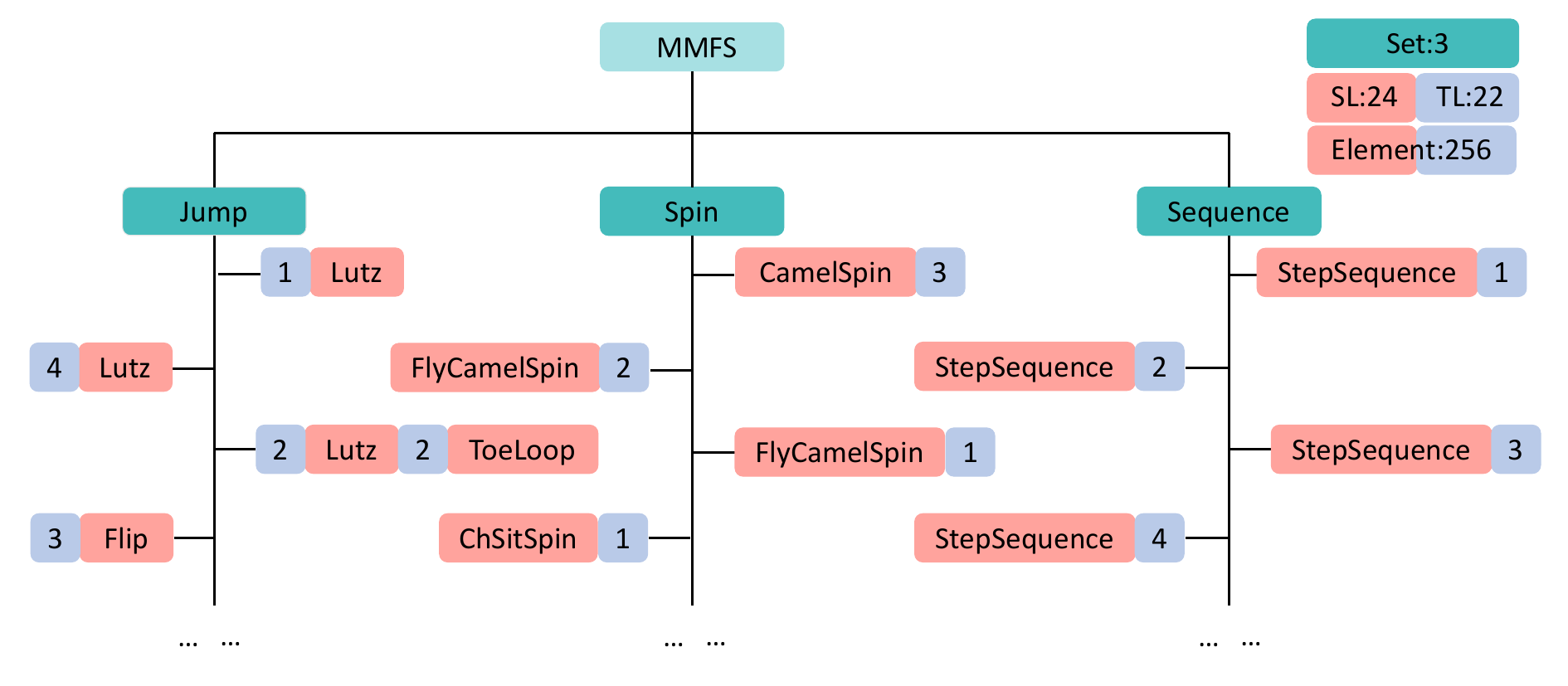}
    \caption{The hierarchical label structure of the MMFS dataset. The actions of each element are fine-grained.}
    \label{fig:fig4}
\end{figure}

\textbf{High Quality.} (1) High Video Quality. All the RGB videos in MMFS are 720p, which benefit describing the subtle difference between clips. High video quality and non-fixed tools are two prerequisites for high-quality videos to extract skeleton features. (2) Strong annotation. Unlike the weak annotation in \cite{sport-1M}, MMFS is strongly annotated on two levels: First, joint annotations are achieved to ensure label reliability by professional annotators combining with the ISU technical document and the provided experts' real-time GroundTruth of the original videos (see Fig. \ref{fig:fig4}). Second, the footage of videos always follows the skater to avoid misclassification due to irrelevant frames.

\textbf{Multi-task.} Generally speaking, action datasets are used for two tasks: action recognition and segmentation. However, Action Quality Assessment \cite{diving48} (AQA) would emerge as an imperative and challengeable issue in MMFS, which can be used to evaluate the action performance of skaters based on BV and GOE scores. As shown in Fig. \ref{fig:fig5}(b), BV and GOE, which depend on action categories and action performance, respectively, are included in our dataset. BV depends on action types and degree of action difficulty. Besides, a 10\% bonus BV score is appended in the latter half of a program.

\textbf{Multi-modality.} We extract the RGB, flow, and skeleton features from the videos in MMFS. Specifically, the skeleton features are obtained using HRNet \cite{cheng2020bottom}(see Fig. \ref{fig:fig4}(b) and more details in supplementary materials). Furthermore, the audio features, which may play important roles in AQA tasks, can also be extracted from videos. Actions matched to musical structure tend to obtain higher GOE scores in the official documentation.

\textbf{Hierarchical Multi-label.} All actions are labeled manually on three levels, coined as set, sub-set, and element. And the sub-set can be divided into the spatial label (SL) and temporal label(TL) as shown in Fig. \ref{fig:fig4}.

\subsection{Dataset Challenge}
For most action recognition datasets, scenes, objects, tools, and persons are essential elements. Many fine-grained actions are generated based on the combination of the person and other elements. MMFS pays more attention to fine-grained action by non-fixed tools (skates). We analyze the fine-grained semantics and the fine-grained complexity in the MMFS, to propose new challenges for the existing models. Figure 3 describes the differences between semantics and complexity. The specific challenges of MMFS are as follows:

\begin{figure}
   \centering
   \includegraphics[width=1\linewidth]{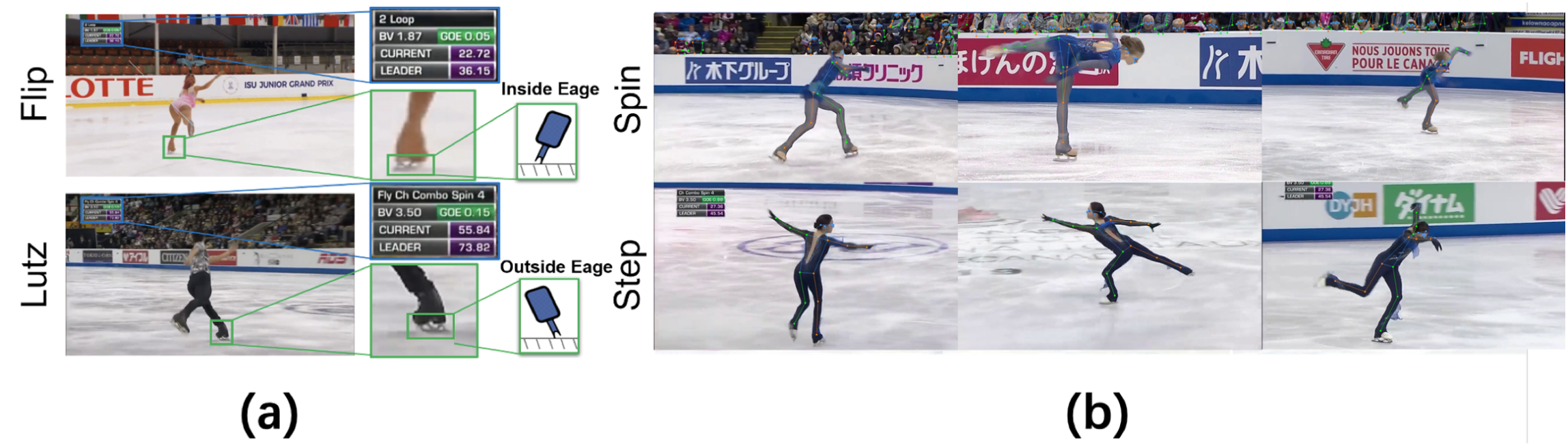}
   \caption{Fine-grained semantics. (a) Misclassification is caused by
subtle spatial variation. (b) Misclassification caused by
partial Spatio-temporal variation. MMFS provides information-board, including BV, GOE, and Groundtruth of classification.}
   \label{fig:fig5}
\end{figure}
\textbf{Fine-grained semantics} The challenges in Fine-grained semantics can be described as the change of labels from the subtle spatio-temporal variation of action units. (1) Temporal variation ($P(cl|tv)\rightarrow1$). It is a problem to determine the number of rotations from a few frames. For example, it is hard to distinguish 2Axel jump and 3Axel jump through limited frames. (2) Spatial variation ($P(cl|pv)\rightarrow1$). It would be difficult to recognize an action by subtle spatial variation of action units. Fig. \ref{fig:fig4}(b) shows the subtle variation between the Flip jump and the Lutz jump. The subtle variation is that the edge of the ice blade is outside on Lutz and inside on Flip. (3) Spatio-temporal variation\cite{RN131} ($P(cl|pv,tv)\rightarrow1$). In Fig. \ref{fig:fig4}(a), the classification will be confused by the similarity features in the partial spatio-temporal variation among classes.

\begin{figure}
   \centering
   \includegraphics[width=1\linewidth]{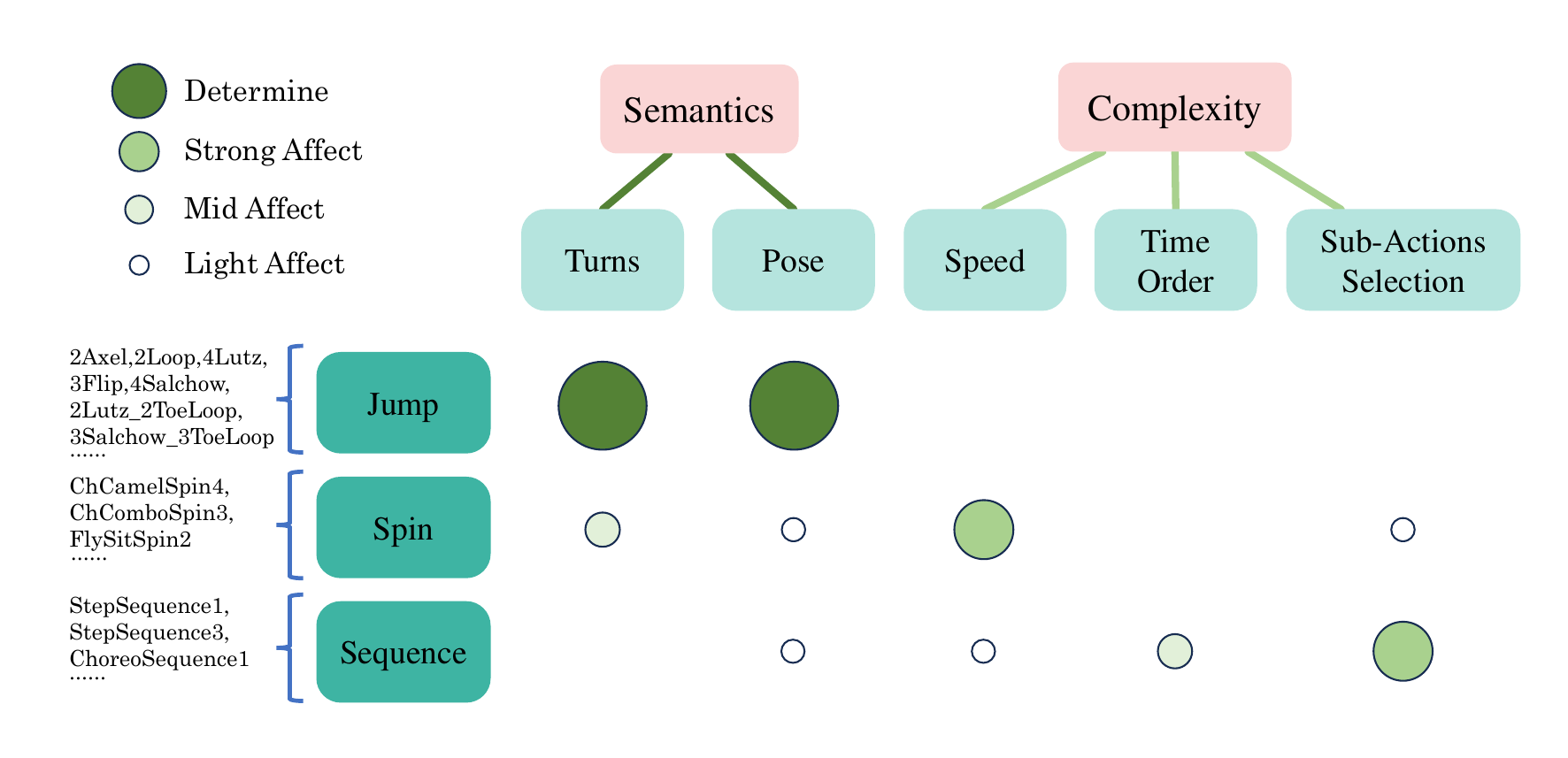}
   \caption{Connections and differences between the fine-grained semantics and the fine-grained complexity. The classification of the Jump set is determined by fine-grained semantics (In fact, the intra-class variance of the jump set will be affected by fine-grained complexity.) while the classification of the Spin set and Sequence set is affected by fine-grained complexity. }
   \label{fig:fig6}
\end{figure}

\begin{figure}
	\centering
	\includegraphics[width=1\linewidth]{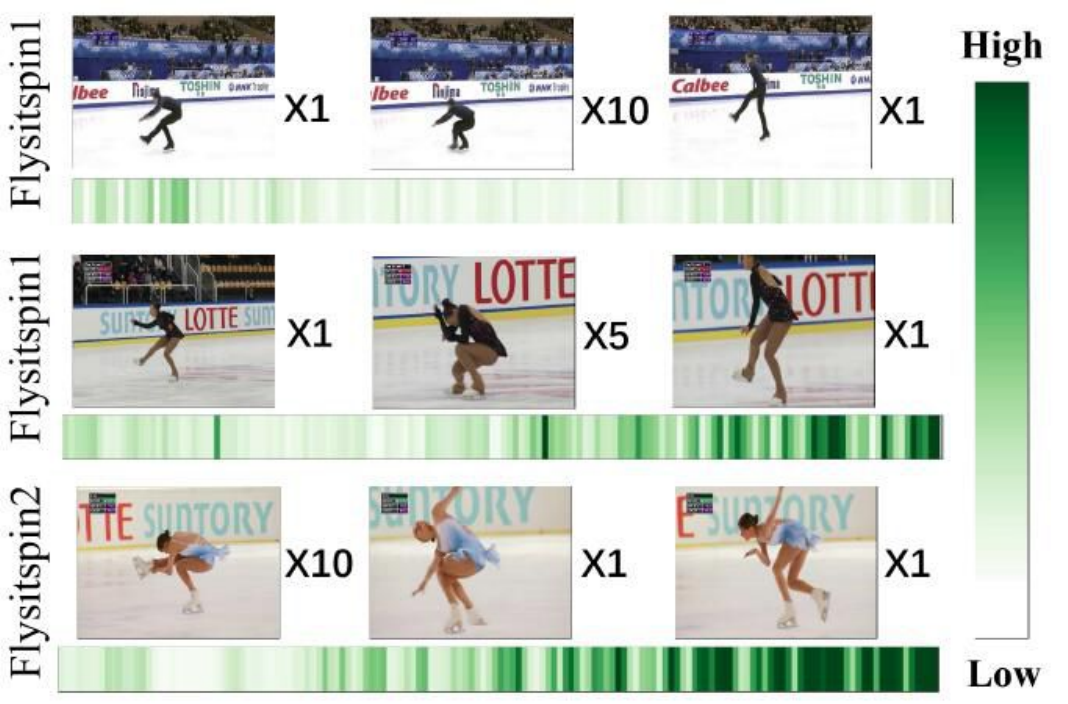}
	\caption{The temporal variation of action units in fine-grained complexity: Seven turns in the middle sample and twelve turns both in the top and the bottom samples.}
	\label{fig:fig7}
\end{figure}

\textbf{Fine-grained complexity} The challenges in Fine-grained complexity are more reflected in the larger inter-class variance and the large duration and speed variance of actions. The detail can be seen in Fig. \ref{fig:fig7}. (1)Temporal variation ($P(cl|tv)\rightarrow0$). The temporal intra-class variance can be demonstrated by the samples in Fig. \ref{fig:fig5}. Although the top two actions belong to the same category, a clear difference in both the action speed and the number of rotations can be detected. Although the two bottom samples in Fig. \ref{fig:fig5} have high similarity in speed, they belong to different actions $P(tv|cl)\rightarrow0$. (2) Spatial variation ($P(cl|pv)\rightarrow0$). The enhanced intra-class variance of action features is mainly reflected by the GOE of actions. The insufficient times of turns and raising hands (Fig. \ref{fig:fig1}) cause GOE deduction and bonus, respectively. More GOE deduction of one action will be caused by hand support, turnover, paralleling feet, and trips during the landing process. Except for GOE, some skaters prefer clockwise rotation while some prefer the opposite. (3) Spatio-temporal variation ($P(cl|pv,tv)\rightarrow0$). The challenge can be demonstrated by the comparison of StepSequence. StepSequence1 requires at least five difficult sub-actions while StepSequence2 requires at least seven difficult sub-actions in the official document. The sub-actions of the same grade StepSequence can be differently combined by a skater.

\section{Experiment}
\label{sec:sec4}

\begin{table}[h]
\begin{center}
\caption{The Top-1 accuracy of RGB-based models tested on MMFS-3 and MMFS-63.}
\label{tab:rgb}
\begin{tabular}{ccc}
\toprule
Method  & MMFS-3 &MMFS-63 \\  \hline

I3D  \cite{RN178}&56.8  & 19.9    \\
TSN \cite{TSN} & 86.5  & 23.4      \\
TSM  \cite{TSM}   & 90.3 &50.9     \\
PAN \cite{PAN} & 91.5 & 69.1 \\

\bottomrule
\end{tabular}
\end{center}

\begin{center}
\caption{The Top-1 accuracy of skeleton-based models tested on MMFS-3 and MMFS-63.}
\label{tab:gugedian}
\begin{tabular}{ccc}
\toprule
Method  & MMFS-3 &MMFS-63 \\  \hline
ST-GCN \cite{yan2018spatial}  &98.9  & 77.4    \\
2S-AGCN \cite{2sagcn} & 99.3  &74.3      \\
CTRGCN \cite{chen2021channel}    & 99.4 &78.8     \\
efficientGCN B4 \cite{song2022constructing} &99.2 &72.1    \\
PoseC3D \cite{PoseC3D} & 99.4 &75.0    \\
\bottomrule
\end{tabular}
\end{center}

\begin{center}
    \centering
    \caption{The performance of RBG-based TSN and skeleton-based PoseC3D pre-trained on MMFS-63, Kinetics and FineGym-99.}
    \label{fig:transfer}
    \setlength{\tabcolsep}{0.1mm}{
    \begin{tabular}{cccc}
    \toprule
    Method                          & MMFS-63            & Kinetics     & FineGym-99  \\
    \midrule
    TSN(no-pre-trained) \cite{TSN}  & \textbf{24.0} & \textbf{70.6} & -\\
    \midrule
    TSN(pre-trained on MMFS-63)         & -              & 62.1  &-         \\
    TSN(pre-trained on Kinetics)    & 22.4          & -      & -        \\
    \midrule
    PoseC3D(no-pre-trained)  \cite{PoseC3D}  & \textbf{77.4} & - & \textbf{93.7} \\
    \midrule
    PoseC3D(pre-trained on MMFS-63)      & -        &-      & 90.1          \\
    PoseC3D(pre-trained on FineGym-99) & 75.8    &-      & -              \\
    \bottomrule
    \end{tabular}}
\end{center}

    \centering
    \begin{center}
    \caption{Action quality assessment on C3D-LSTM, C3D-AVG-MTL, CoRe and DAE-MLP.}
    \label{tab:AQA}
    \begin{tabular}{cc}
    \toprule
              Methods      & SC    \\ \hline
    C3D-LSTM   \cite{parmar2017learning}    & 0.5234   \\
    C3D-AVG-MTL  \cite{MTL-AQA}     & 0.3831    \\
    CoRe	\cite{yu2021group}	& 0.7313 \\
    DAE-MLP \cite{zhang2021auto} 	& 0.5915\\ \bottomrule
    \end{tabular}
    \end{center}
\end{table}

\subsection{Experimental Preparation}
In MMFS-63, all the samples are divided into 4113 and 991 clips for training and testing. Set-level of MMFS, sub-set-level (Temporal Label 22 (TL22) and Spatial Label 24 (SL24)). And fine-grained elements-level (MMFS-63) are annotated by the different semantic labels. We use 30 fps of RGB videos and extract skeleton features with 17 joints for each frame by leveraging HRNet in MMFS-63.

To better understand the performance of prominent action recognition models on this proposed dataset, we benchmark a variety of models on MMFS and group the models into two categories: RGB-based models and skeleton-based models.

\textbf{RGB-based Models.} For RGB-based action recognition, models process very high dimensional input and are more sensitive to the size of training data. Several prominent action recognition models are selected as test methods. Specifically, the RGB-based experiments are conducted utilizing I3D \cite{RN178}, TSN \cite{TSN}, TSM \cite{TSM}, and PAN\cite{PAN} methods. As for action quality assessment, C3D-LSTM \cite{parmar2017learning}, C3D-AVG-MTL \cite{MTL-AQA}, CoRe \cite{yu2021group}, and DAE-MLP \cite{zhang2021auto} are utilized for the baseline methods.

\textbf{Skeleton-based Models.} We adopt the skeleton-based models on this dataset, including ST-GCN \cite{yan2018spatial}, 2S-AGCN \cite{2sagcn}, CTRGCN \cite{chen2021channel}, efficientGCN B4 \cite{song2022constructing}, and PoseC3D\cite{PoseC3D}. For the skeleton-based methods, the large duration variance of clips (the length range of clips is between 25 and 2536 frames) motivates us to use the average frame number of all clips (320 frames) to construct the input\footnote{The 320 frames are extracted from equal divisions of each clip. The clip with insufficient frames (less than 320 frames) should be padded by zeros instead of skeleton features}.

In the benchmark, we focus on fine-grained action recognition with multi-modality, spatial and temporal semantics comparison, and the performance of mainstream methods in action quality assessment. The parameterization of all models can be found in the supplemental material.
\subsection{Fine-grained Action Recognition and Quality Assessment}
\textbf{Multi-modality Action Recognition.} For image-based videos, RGB modality is utilized to extract the spatial content of frames while the skeleton modality could extract the full-body motion features, which have removed most spatial appearance contents. In MMFS, the accuracies of skeleton modality in Tab. \ref{tab:gugedian} are substantially enhanced compared with the results of RGB-based modality in Tab. \ref{tab:rgb}. The results of Tab. \ref{tab:rgb} and Tab. \ref{tab:gugedian} illustrate that MMFS is more discriminative in motion feature variation of body pose and is not sensitive to the visual scene.

\textbf{The Comparison of the Action Quality Assessment task.} For action quality assessment, we adopt the Spearman correlation coefficient (SC) as the metric of experiments. As shown in Tab. \ref{tab:AQA}, the mainstream method has achieved effective but not excellent accuracy on our dataset, which shows that our dataset can bring new challenges to the evaluation task.

\subsection{The Comparison of Spatial and Temporal Semantics}
\textbf{Hierarchical Label.} Different from the coarse-grained dataset, 3 sets in MMFS are divided into 63 action categories to propose a fine-grained action dataset. As shown in Tab. \ref{tab:rgb} and Tab. \ref{tab:gugedian}, the performance of all the compared models drops a lot when the fine granularity is considered on MMFS. The three sets can not achieve outstanding performance with TSN \cite{TSN}, while ST-GCN \cite{yan2018spatial} presents better results based on the features of 320 frames. However, the performance of ST-GCN \cite{yan2018spatial} is also limited to the Spin and Sequence sets. We show the confusing actions in the supplemental material. And the most confusing actions are the Spin set, where more fine-grained temporal semantics will be addressed because of the longer length of duration.

\textbf{The Comparison over SL and TL.} To observe which one occupies more important influence in fine-grained recognition between spatial semantics and temporal semantics, we propose TL22 and SL24 on the sub-set level. As shown in Tab. \ref{tab:TA and SA}, the action recognition accuracy of temporal label division (TL22) achieves worse performance than that of spatial division (SL24). It illustrates that temporal action recognition is more challenging than the same task in the spatial division. The similar recognition results on TL22 and MMFS-63 demonstrate that most of the difficulties focus on the temporal action recognition task. The experimental results above demonstrate that the existing action recognition models fail to extract temporal discriminant features on both the skeleton and RGB-based modalities.

\begin{figure}
  \centering
  \includegraphics[width=1.0\linewidth]{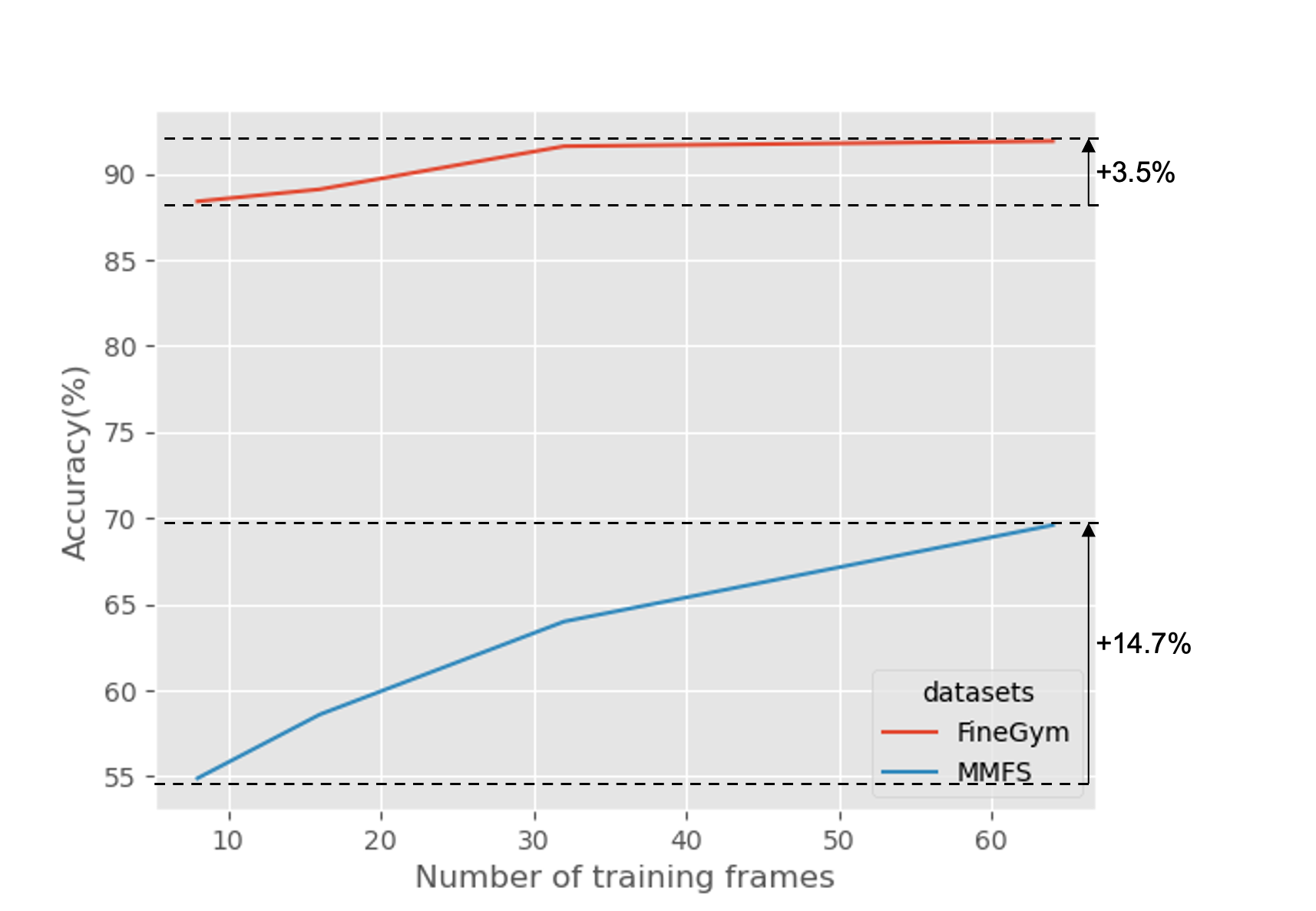}
  \caption{The accuracy of frame extraction on FineGym and MMFS-63 using CTR-GCN. }
  \label{fig:figure9}
\end{figure}

\textbf{The Key Challenge in Temporal Semantics.} As shown in Fig. \ref{fig:figure9}, with the increase in the number of selected frames, CTR-GCN can achieve significant growth on our data set, while FineGym99 has only achieved a small increase. This shows that despite the fine-grained datasets are more sensitive to temporal variance, the temporal feature is difficult to be extracted on our MMFS dataset.

\begin{table}[]
\begin{center}
    \caption{Fine-grained action recognition on spatial and temporal semantics.}
    \label{tab:TA and SA}
\centering
    \setlength{\tabcolsep}{0.8mm}{
    \begin{tabular}{c|cc|cc}
    \toprule
                          & \multicolumn{2}{c|}{Skeleton} & \multicolumn{2}{c}{RGB (16 frames)} \\
    \multicolumn{1}{l|}{} & CTRGCN \cite{chen2021channel}        & PoseC3D \cite{PoseC3D}       & TSM \cite{TSM}             & PAN \cite{PAN} \\ \hline
    MMFS-63               & 78.8          & 75.0         & 50.9              & 69.1              \\
    TL22                  & 76.7          & 78.4         & 51.2             & 71.3              \\
    SL24                  & 92.2         & 95.4        & 85.1             & 91.5              \\ \bottomrule
    \end{tabular}}
    \end{center}

\end{table}

\section{Conclusion}

In this paper, we propose a Multi-modality and Multi-task Dataset of Figure Skating (MMFS) to further research on fine-grained analysis. Distinguishing from the existing fine-grained action datasets, MMFS contains more fine-grained semantics including spatial semantics and temporal semantics. All 11671 clips are annotated with a hierarchically multi-label structure and fine-grained analysis can be conducted on multi-modality. We evaluate the mainstream methods based on RGB-based models and skeleton-based models. In our experiments, we highlight that temporal semantics is more difficult and complex than spatial semantics for the existing models and the skeleton modality achieves better performance on fine-grained analysis.

{\small

\bibliographystyle{ieee}
\bibliography{egbib}
}

\newpage

\vfill

\end{document}